\documentclass{article}
\usepackage{spconf,amsmath,graphicx,hyperref}
\usepackage{threeparttable}
\usepackage{pifont}
\usepackage{multirow}
\usepackage{listings}
\usepackage{xcolor}

\usepackage{subcaption}
\usepackage{amsmath,amsfonts}
\usepackage{array}
\usepackage{textcomp}

\usepackage{mathtools}
\usepackage{booktabs}
\usepackage{wrapfig}
\definecolor{codegreen}{rgb}{0,0.6,0}
\definecolor{codegray}{rgb}{0.5,0.5,0.5}
\definecolor{codepurple}{rgb}{0.58,0,0.82}
\definecolor{backcolour}{rgb}{0.95,0.95,0.92}

\lstdefinestyle{mystyle}{
    backgroundcolor=\color{backcolour},   
    commentstyle=\color{codegreen},
    keywordstyle=\color{magenta},
    numberstyle=\tiny\color{codegray},
    stringstyle=\color{codepurple},
    basicstyle=\ttfamily\footnotesize,
    breakatwhitespace=false,         
    breaklines=true,                 
    captionpos=b,                    
    keepspaces=true,                 
    numbers=left,                    
    numbersep=5pt,                  
    showspaces=false,                
    showstringspaces=false,
    showtabs=false,                  
    tabsize=2
}

\title{DetailCLIP: Injecting Image Details into CLIP's Feature Space}

\name{
{\fontsize{10pt}{10pt}\selectfont
  Zilun Zhang$^{1,\ast}$,
  Cuifeng Shen$^{2,\ast}$,
  Yuan Shen$^{3}$,
  Xinyu Zhou$^{3}$,
  Huixin Xiong$^{3}$,
  Tiancheng Zhao$^{4,5\dagger}$,
  Jianwei Yin$^{1,\dagger}$%
  \thanks{$^{\ast}$Equal contribution. $^{\dagger}$Corresponding authors. \\  \textbf{Email:} \href{mailto:tianchez@zju-bj.com}{tianchez@zju-bj.com}, \href{mailto:zilun.zhang@zju.edu.cn}{zilun.zhang@zju.edu.cn}
}
}
}
\address{
{\fontsize{10pt}{10pt}\selectfont
  $^{1}$Zhejiang University \ 
  $^{2}$Peking University \ 
  $^{3}$Megvii \
  $^{4}$Om AI Research \
  $^{5}$Binjiang Research Institute of Zhejiang University\
}
}

\begin{document}
\ninept
\maketitle
\begin{abstract}
Although CLIP-like Visual Language Models provide a functional joint feature space for image and text, due to the limitation of the CILP-like model's image input size (e.g., 224), subtle details are lost in the feature representation if we input high-resolution images (e.g., 2240). Our proposed framework addresses this issue by generating a \textbf{single feature representation} for a high-resolution image that retains image details from different scales while sharing the same semantic space as the original CLIP. \textcolor{black}{An application scenario is remote sensing text–image retrieval, where targets (e.g., vehicles and ships) often appear at tiny scales.}
To achieve this, we develop a feature fusion model that relies on CLIP features extracted from a carefully designed image patch method, dubbed Complete Cover. This method ensures comprehensive coverage of objects across various scales and is weakly supervised by image-agnostic class prompted queries. We evaluate our framework's performance using real-world and synthetic datasets, demonstrating significant improvements in image retrieval tasks based on class prompted queries. To further showcase our framework's capability in detail retrieval, we introduce a CLEVR-like synthetic dataset, named CLVER-DS. This fully annotated dataset offers a controllable object scale, allowing for a more thorough examination of our approach's effectiveness. Our code is publicly available at https://github.com/zilunzhang/DetailCLIP
\end{abstract}

\section{Introduction}

\label{sec:intro}

\begin{figure}[ht]
    \centering
    \includegraphics[width=0.5\textwidth]{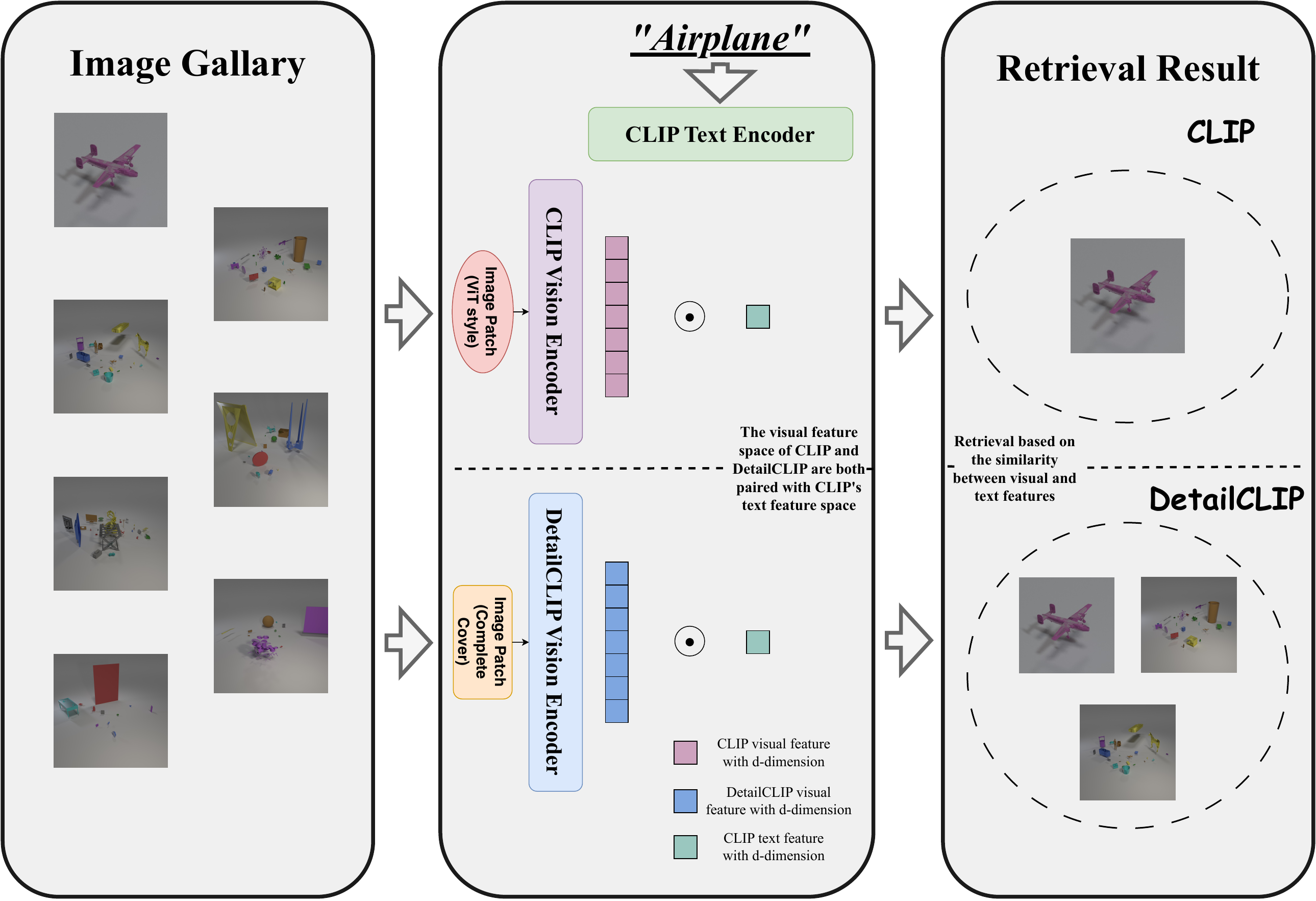}
    \caption{CLIP vs. DetailCLIP. Given the "Airplane" query, vanilla CLIP retrieves only the large target. In contrast, DetailCLIP retrieves all relevant images, including those with extremely small targets. Both models share the text encoder (light green). CLIP's vision encoder is light purple; DetailCLIP's is light blue. The symbol of a dot within a circle denotes the dot product for cosine similarity.}
    
    \label{fig:title_image}
\end{figure}

Text-to-image retrieval task involves retrieving pertinent images based on a text query, which could either describe the entire image or focus on a specific and small object within it. For example, when using CLIP to search for images containing a red helmet with "red helmet" as the text query, the top results typically showcase a large red helmet at the center of the image. However, we might also be interested in images of people wearing red helmets on a football field, where the helmets appear much smaller. Since CLIP is designed to match an entire image to a text description, it struggles to retrieve such images using only the "red helmet" query. \textcolor{black}{Moreover, for remote sensing text-to-image retrieval, where users search massive satellite archives with short queries (e.g., “ship/airplane/vehicle”), small targets are easily lost in global representations}. These limitations pose a challenge when attempting to retrieve all related images in a database using a single word, as illustrated in Figure \ref{fig:title_image}.

Existing solutions are suboptimal: patch-based retrieval features are not useful for downstream tasks; detector-based methods (e.g., Crop-CLIP \cite{crop-clip} and RegionCLIP \cite{regionclip}) are limited by the detector's predefined classes; and other fine-grained models (e.g., MDETR \cite{kamath_mdetr_2021}, XDETR \cite{cai2022x} and FILIP \cite{yao2021filip}) require massive retraining data and computational resources. To address this, we propose Detail Injected CLIP (\textbf{DetailCLIP}), a novel framework that efficiently captures fine-grained details at minimal cost.
Our approach first identifies the object scales where CLIP performs effectively. We leverage this insight to design "Complete Cover (\textbf{CC})", an image patching scheme that contains objects in different scales. A transformer then fuses the CLIP features extracted from these CC patches, and a self-supervised learning loss aligns features with detailed information into this newly fused single-feature vector. \textbf{DetailCLIP} \textbf{integrates detailed image information into a single vector} suitable for end-to-end training, requiring only \textbf{the cost of training a tiny fusion model}. 
\textcolor{black}{It supports small-target text-to-image retrieval of high-resolution imagery (e.g. remote sensing) by preserving multi-scale details while producing a single indexable feature.}

Our main contributions are: (1) We propose \textbf{DetailCLIP}, an efficient framework that generates a single, detail-rich image feature, unlike CLIP's global-only representation. It shows superior performance on MSCOCO, LVIS, and synthetic datasets and can serve as a \textbf{plug-in module} to improve detail retention \textbf{for other Vision-Language Models}. (2) We introduce "Complete Cover (\textbf{CC})", a novel patching scheme that covers all object scales using a pre-defined iterative process. It \textbf{requires no fine-grained annotations} (e.g., bounding boxes), needing \textbf{only dataset-level class names} for training. (3)We develop "CLEVR of Different Scales (\textbf{CLEVR-DS})", a new fully annotated, scale-controllable retrieval benchmark using CLEVR \cite{johnson2017clevr} and ShapeNet \cite{ShapeNet2015}, where our method significantly outperforms existing work.

The rest of this paper is organized as follows: Section \ref{relatedwork} discusses related works; Section \ref{method} provides a detailed description of our proposed "Complete Cover (\textbf{CC})" method and the \textbf{DetailCLIP} framework; SectionSection \ref{benchmark} introduces our proposed benchmark; and Section \ref{experiment} presents experiments that validate the superiority of our framework.

\section{Related Work}
\label{relatedwork}

\subsection{Overview of Vision-Language Models} 
Pre-trained Vision-Language Models (VLMs) are often classified by objectives like image-text contrastive learning (ITC), matching (ITM), or masked modeling (MLM) \cite{du2022survey}. ITC models like CLIP \cite{radford_learning_2021} and ALIGN \cite{jia_scaling_2021} showed robust adaptability using massive, noisy image-text pairs. Variants improved this via detailed token alignment (FILIP) or enhanced supervision (SLIP \cite{slip}). ITM-based models (e.g. ViLBert \cite{vilbert}, ALBEF \cite{albef}) focus on fine-grained representation alignment, while MLM-based models (e.g., FLIP \cite{flip}, BEIT3 \cite{beit3}) proved effective. Innovations like in-context learning (Flamingo \cite{alayrac2022flamingo}), captioning (CoCa \cite{coca}), and Mixture of Experts (VLMo \cite{vlmo}) have been introduced to further refine the capabilities of pre-trained VLMs. Many models mix objectives, such as ALBEF (ITC+ITM) or FLIP (ITC+MAE). Recent models like InternVL-C \cite{chen2024internvlscalingvisionfoundation} also show powerful performance in cross-modal retrieval. While CLIP-like models show strong zero-shot robustness \cite{goh2021multimodal}, they are prompt-sensitive \cite{zhou2021learning} and lack fine-grained alignment. Improvement include adding image self-supervision (SLIP) or new training objectives (DeCLIP). Advanced models like GLIP \cite{glip} and MM Grounding-DINO \cite{zhao2024opencomprehensivepipelineunified} unify detection and phrase grounding for object-level understanding.

\subsection{Applications of Vision-Language Models}
VLMs are widely applied to downstream tasks. MDETR and XDETR \cite{xdetr} advanced multi-modal detection. RegionCLIP generates region pseudo-labels from CLIP for open-vocabulary detection. GroupViT \cite{groupvit} learns to group semantic regions for segmentation using only text supervision. FlexiViT \cite{flexivit} randomizes patch size during training to achieve robustness to varied patch sizes at inference. RegionSpot \cite{regionspot} learns position-aware localization by combining a localization foundation model with a semantic information extraction model.

\section{Methodology} \label{method}

\subsection{Motivation and Effective Scale Sensitivity}

\label{sec:motivation_ess}
We verify CLIP's declining retrieval performance on small objects using the LVIS dataset \cite{gupta2019lvis}, which is richer in small object annotations and categories than COCO \cite{lin2014microsoft}. To analyze this, we create LVIS subsets by varying the upper bound on the maximum relative object area, $r_\text{max}$, defined as:
\begin{equation}
     r_{\text{max}} = \frac{\text{Maximum Area of the Objects in the image}}{\text{Area of the Image}} 
\end{equation}
\noindent As shown in Table \ref{lvis_stats}, CLIP's performance consistently degrades as this upper bound $r_\text{max}$ (and thus the maximum object size) is reduced.



\begin{table}[bpht]
  \centering
  \begin{tabular}{cccc}
    \toprule
    $r_\text{max}$ & \tnote{$\dag$}\,\, Recall@1 & Recall@3 & Recall@5 \\ 
    \midrule
        $10^{0}$ & 8.63\% & 15.19\% & 18.60\%  \\ 
        $10^{-0.5}$ & 7.52\% & 13.87\% & 17.45 \% \\ 
        $10^{-1}$ & 5.75\% & 11.28\% & 14.66 \%  \\ 
        $10^{-1.5}$ & 4.92\% & 9.61\% & 12.16 \%  \\ 
        $10^{-2}$ & 3.98\% & 8.76\% & 11.48 \%  \\ 
  \bottomrule
\end{tabular}
\begin{tablenotes}
       \item[$\dag$] Refer to section \ref{Evaluation Metrics} for the metric detail
\end{tablenotes}
\caption{Performance of CLIP text-image retrieval task with different LVIS subsets.}

  \label{lvis_stats}
\end{table}

We define the \textbf{Effective Scale Sensitivity (ESS) of CLIP-like models as the minimum object occupying percentage in an image that CLIP-like models can retrieve}. We need to input images within the sensitivity of CLIP-like models. A natural thought is slicing an image into small patches and retrieving objects on those patches. Consider the side length of an image is $n$, and the number of possible patches to cover all possible objects is $O(n^4)$ level which is unbearable. We propose the "Complete Cover (\textbf{CC})" method to eliminate redundant patches while covering all possible objects.

\subsection{Problem Definition}

In this section, we provide a formal problem definition for Detail Injection using CLIP-like models, an image, and a set of image patches. Suppose we generate $p$ patches from an image $\mathbf{X}$, and we denote the set of image patches as $x_{i} \in \mathbf{X}$, where $i \in {1, ..., p}$. Then, $\mathcal{F}: \mathbb{R}^{c \times w \times h} \xrightarrow[]{} \mathbb{R}^{d}$ represents the CLIP-like models' image encoder, and the $d$-dimensional feature $u_i$ extracted from a single image patch $x_i$ can be expressed as:

\begin{equation}
  u_{i} = \mathcal{F}(x_{i}),
\end{equation}

We denote the set of image patch features for a given image as $\mathbf{U} = \{u_{i}\}$, where $i \in \{1, ..., p\}$. We define our fusing model as $\mathcal{D}: \mathbb{R}^{p \times d } \xrightarrow[]{} \mathbb{R}^{d }$. The DetailCLIP feature $v$ is obtained from:

 \begin{equation}
      v = \mathcal{D}(\mathbf{U}),\,\text{where}\,v \in \mathbb{R}^{d}
 \end{equation}

\subsection{Complete Cover} 
\label{completecover}

\begin{figure}[htbp]
    \centering
    \includegraphics[width=0.35\textwidth]{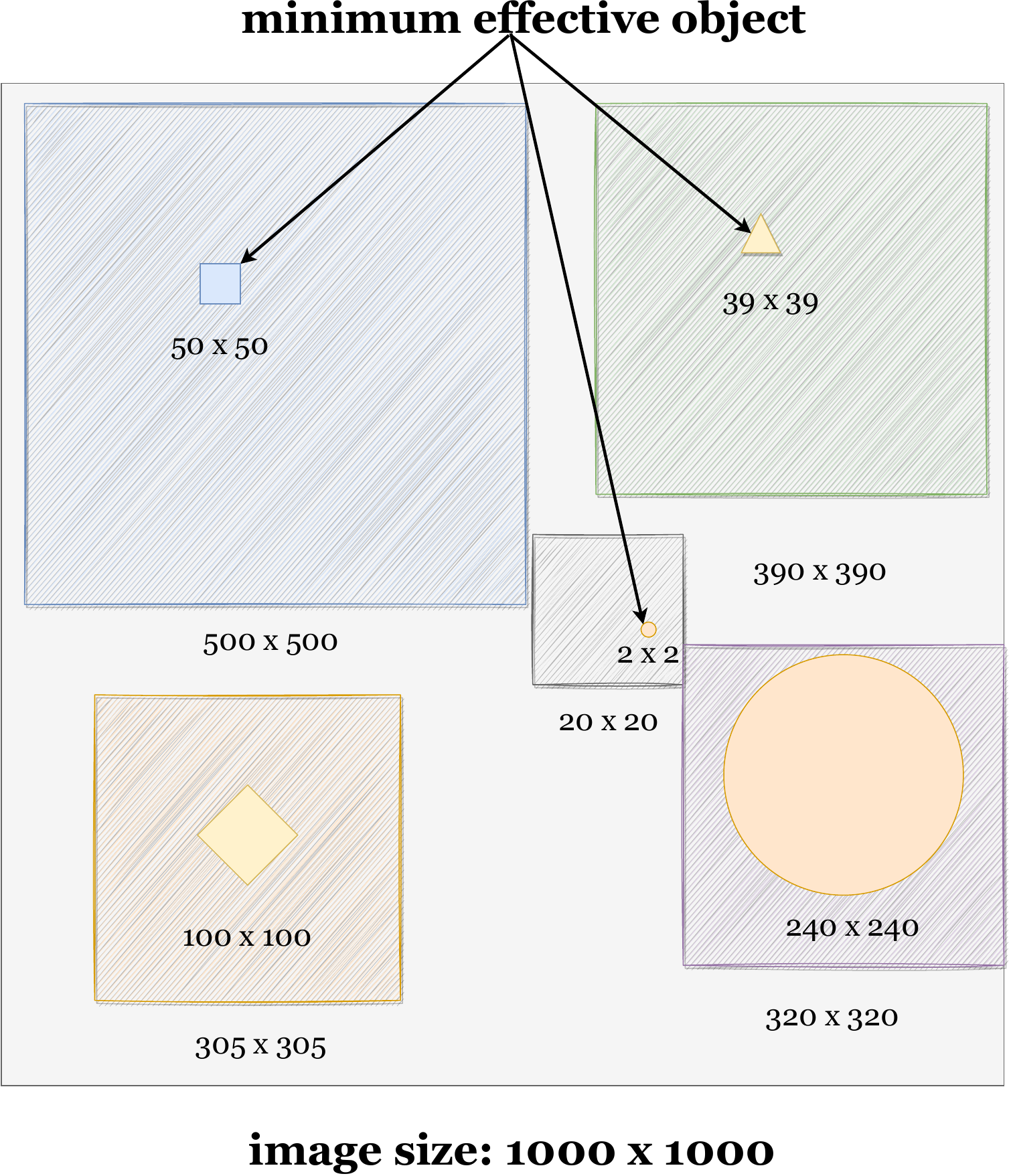}
    \caption{Illustration of Complete Cover (CC). Minimum effective object means the minimum object that can be retrieved by CLIP from the patch. Patches with different sizes will slide the whole image to cover objects equal to or bigger than the minimum effective size.}
    \label{fig:patchcc_left}
\end{figure}

In this subsection, we introduce a patch generation scheme. Given an image with a side length of $n$, the complexity of the number of possible patches needed to cover all objects is on the order of $O(n^4)$, which is computationally infeasible. To address this issue, we propose the "Complete Cover (\textbf{CC})" method, which aims to eliminate redundant patches while ensuring complete coverage of all possible objects. A schematic representation of the \textbf{Complete Cover} method can be found in Figure \ref{fig:patchcc_left}.

\subsubsection{Algorithm}

First, we define the concept of \textbf{cover}. Let $Q$ represent the set of pixels in a patch, and $P$ denote the set of pixels within an object's bounding box.
We denote “c” as the effective scale sensitivity defined in section \ref{sec:motivation_ess} and "a" to be the side length of $P_0$ introduced later.
To ensure that the patch includes objects while retaining its retrieval capability, we define $Q$ covers $P$: 
\begin{equation}
    \label{cover}
    \begin{aligned}
    Q\,\text{covers}\,P
    &\coloneqq C(Q,P) \\
    &= 
    \begin{cases}
    \text{1}, & \forall p = (x, y) \in P \rightarrow p \in Q\,\text{and}\,|P| > c^2\cdot|Q|\\
    \text{0}, & \text{Otherwise}
    \end{cases} \\
    &\text{where} \,|\cdot| \, \text{is the number of pixels in}\,\cdot \, \text{, and c is} \\
    & \text{effective scale sensitivity defined in our paper}\hfill
    \end{aligned}
\end{equation}
Next, we define the set of all possible $P$ in an image as $S_\text{full}$. In the "Complete Cover (\textbf{CC})" scheme, we design a greedy algorithm to generate a set of $Q$ as $S_\text{cc}$, which fulfills:

\begin{equation}
\forall\,P \in S_{\text {full}}, \exists\,Q \in S_{\text {cc}},s.t.\, C(Q, P)=1
\end{equation}


The specific patch selection method proceeds as follows:
Given an effective scale sensitivity $c$, assume the full image pixel set is $Q_0 \in S_\text{cc}$, and all object bounding box pixel sets $P_0 \in S_\text{full}$ that it can cover satisfy:

\begin{equation}
C(Q_0, P_0) =1, \forall P_0 \in S_\text{full} ,\text{if}\,|P_0| > c^2\cdot|Q_0|,
\end{equation}

\begin{figure}[htbp]
    \centering
    \includegraphics[width=0.4\textwidth]{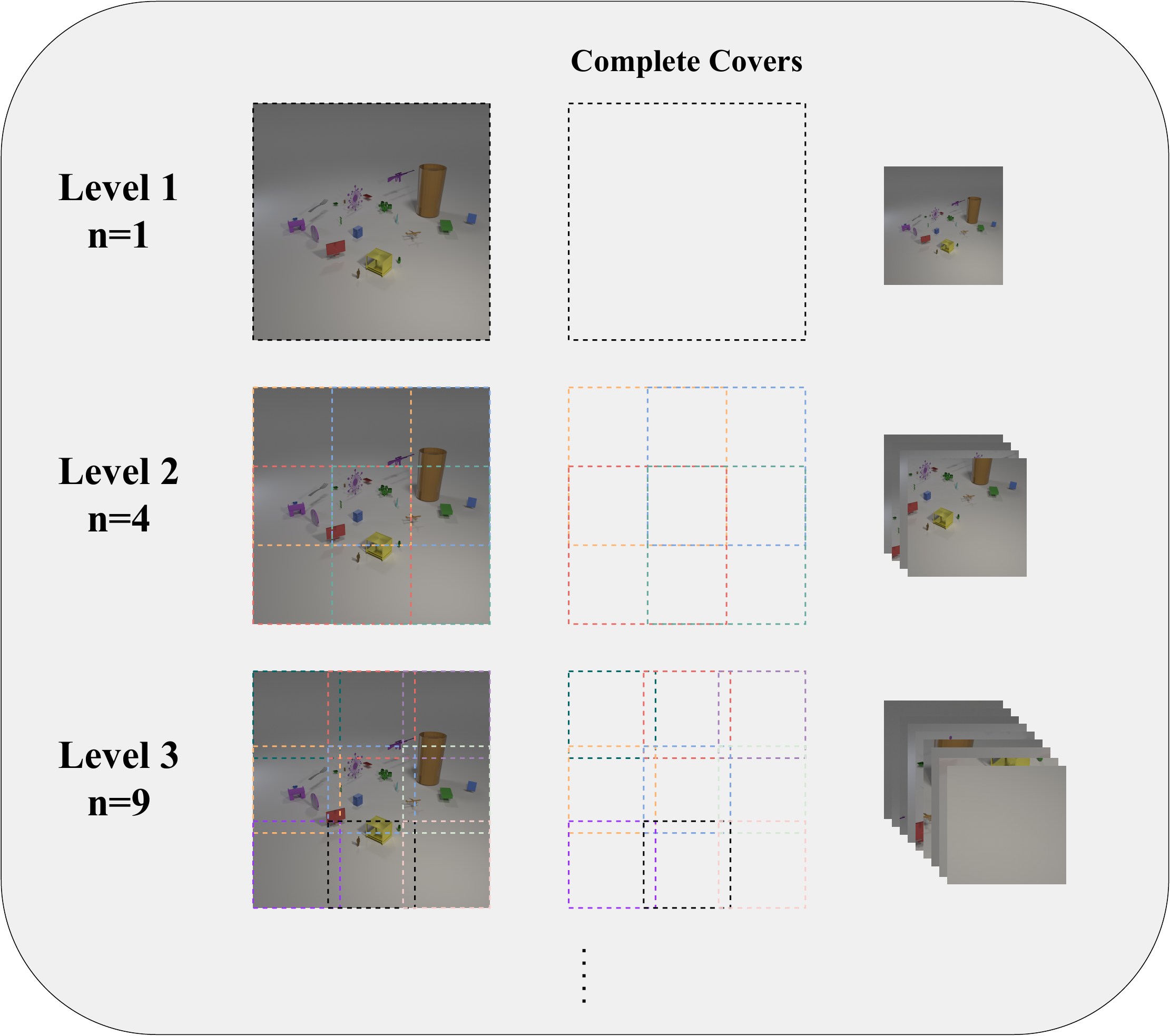}
    \caption{A diagram shows the first three levels of the CC scheme for $\frac{a}{c}=10$. Each color represents a patch.}
    \label{fig:patchcc_right}
\end{figure}

Without loss of generality, let $P\in S_{\text{full}}$ be square and $P_0$ have a side length of $a$. In order to cover $P_1 \in S_{\text{full}}$ with a side length of $a/c - 1$, we employ a greedy approach to obtain $Q_1 \in S_{\text{cc}}$ such that $C(Q_1, P_1)=1$. This is achieved by passing a global sliding window with a side length of $a-c$ and a step size of $a/c - 2$. We repeat this procedure until we have patches that can cover objects with side lengths ranging from $a/c$ to $a/c -n$, where $n=a/c-1$. The complexity of the number of possible patches needed to cover all objects reduces from $O(n^4)$ to $O(n^2)$.
While regular sliding windows can act as covers, they may not be “complete” covers, as they might not fully cover all possible objects. Figure \ref{fig:patchcc_right} showing the first three levels of the complete cover scheme for $\frac{a}{c}=10$.


\subsubsection{The Effectiveness of Complete Cover} \label{cc_prove}
Let us explore the computational complexity of the number of all possible patches using a brute force algorithm. For an image with a side length of $a$, the total number of possible patches is $O(a^4)$, as each rectangular patch can be defined by its top-left and bottom-right corner coordinates. When limiting the patches to be square, the complexity remains significant at $O(a^3)$. 

To simplify our analysis, we focus on square patches. In our main paper, we define \textit{$c$} as the ratio of the perimeter, which is equivalent to the ratio of the side length. By adopting the Complete Cover scheme, we can generate patches at different levels with sidelengths of $[c, 2c, 3c, \cdots, a]$. For each level, the number of patches to cover all targets for corresponding side length is $O( \left(\frac{a}{c}\right)^{2}), O( \left(\frac{a}{2c}\right)^{2}), O( \left(\frac{a}{3c}\right)^{2}), \cdots , O( \left(\frac{a}{a}\right)^{2} )$ respectively. 

The total number of patches introduced by Complete Cover is:

\begin{align}
     & \left(\dfrac{a}{c}\right)^{2}  * (1 + \dfrac{1}{2^2} + \dfrac{1}{3^2} + \cdots ) \\
    =& \left(\dfrac{a}{c}\right)^{2}  \times \dfrac{\pi^2}{6} \\
    =& O(a^2)
\end{align}
where $c$ is a constant across the experiment.

\begin{figure}[htbp]
    \centering
    \includegraphics[width=0.45\textwidth]{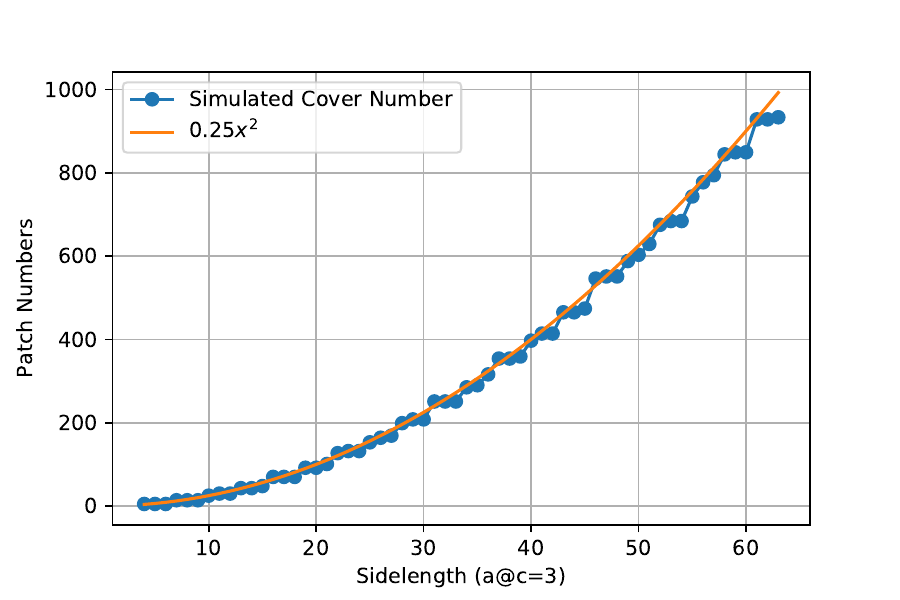}
    \caption{Patch numbers for different sidelengths when $c = 3$.}
    \label{fig:simu_cc}
\end{figure}

In Figure \ref{fig:simu_cc}, we illustrate the number of patches for various side lengths when $c=3$, alongside a quadratic function $y=0.25x^2$. The close fit between these two curves highlights the effectiveness of our approach.

It is essential to note that before cropping patches, we resize the image to ensure the side length $a$ is divisible by $c$. The minimum effective range for Q is determined as the range of side lengths for P that results in $C(P, Q)=1$. We present the formulas for cover side length, minimum effective range, and the patch number for a given $c$ in Table \ref{table:general_term}.

\begin{table}[ht]

    \begin{center}
    \renewcommand{\arraystretch}{1.6}
    \resizebox{0.98\columnwidth}{!}{
    \begin{tabular}{|c|c|c|c|}
    \hline
        Level & Cover Sidelength & Minimum Effective Range & Patch Numbers \\ \hline
        1 &  $a$ & $a \geq x \geq \frac{a}{c} $ & 1 \\ \hline
        2 & $a - c$ &  $\frac{a}{c}  > x \geq \frac{a-c}{c}  $ & $(\frac{2c + ac - a}{ -c^2 +2c + ac - a} ) ^ 2$ \\ \hline
        3 &   $a - 2 c$ & $\frac{a - c}{c}  > x \geq \frac{a-2c}{c} $ & $(\frac{3c + ac - a}{ -2c ^ 2 + 3c + ac - a} ) ^ 2$ \\ \hline
        $\cdots$ & $\cdots$ & $\cdots$ & $\cdots$ \\ \hline
        n &$(\frac{nc + ac - a}{ -(n-1)c ^ 2 + nc + ac - a} ) ^ 2$ & $\frac{a - (n-2) c}{c} > x \geq \frac{a - (n-1)c}{c} $ & $a - (n-1) c$  \\ \hline
        $\cdots$ & $\cdots$ & $\cdots$ & $\cdots$ \\ \hline
    \end{tabular}
    }
    \end{center}
    \caption{Relationship between the cover sidelength, the minimum effective range, and the number of patches at a given c}
    \label{table:general_term}
\end{table}

Our \textbf{CC} method can better retain detailed information compared to simply slicing the image into non-overlapping, equal-sized patches as proposed by ViT \cite{vit}. \textbf{CC} faces a trade-off between completeness and computational complexity with different values of $c$. A reasonable choice of $c$ should generate a manageable number of patches while ensuring the preservation of detailed information.

\subsection{Model \& Loss} \label{modelloss}

\begin{figure}[ht]
    \begin{center}
    \includegraphics[width=0.5\textwidth]{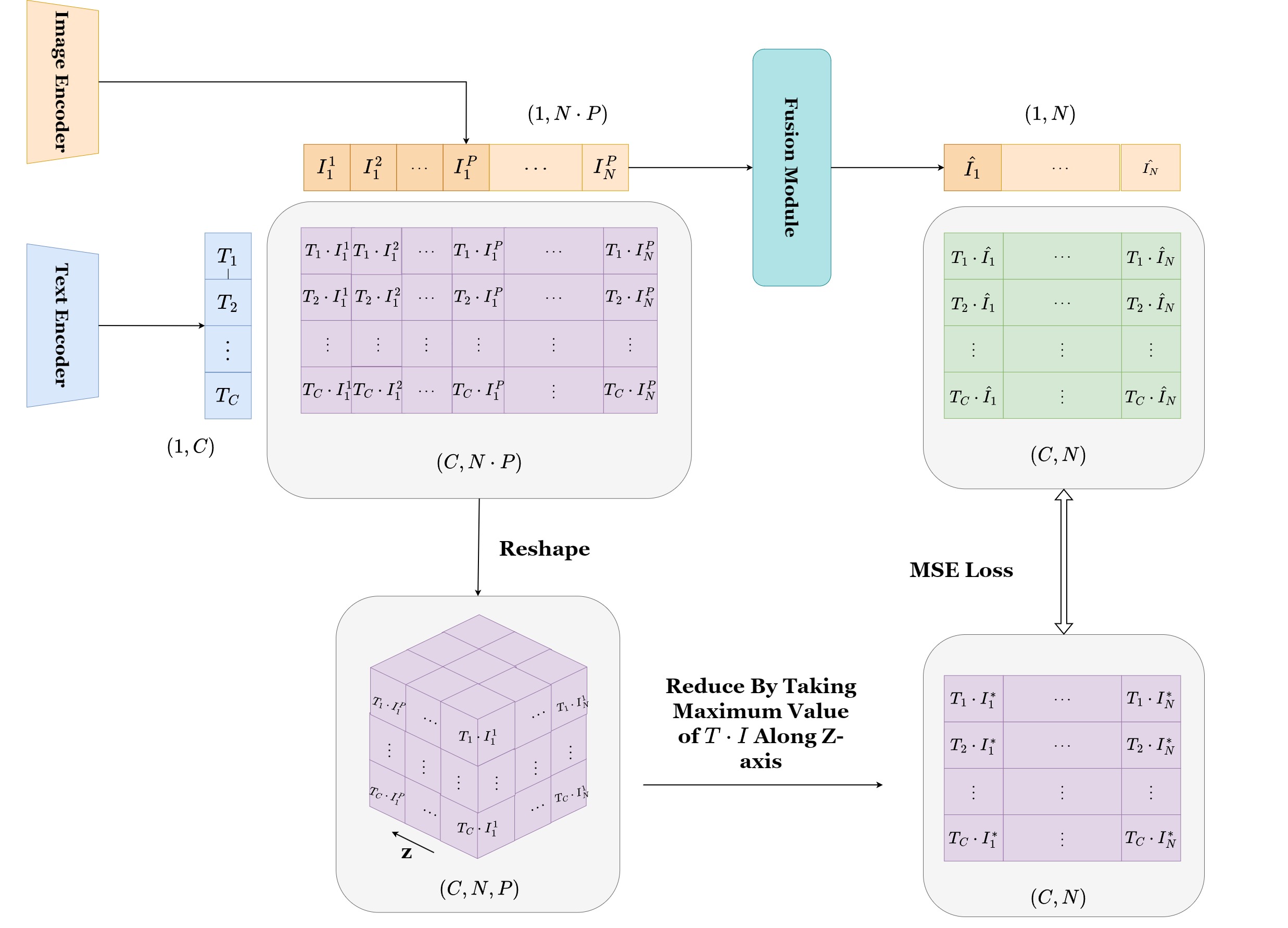}
    \caption{The DetailCLIP framework. 
    Here, $N$, $P$, and $C$ are the number of images, patches per image, classes, respectively. $I$, and $T$ are image and text features.
    $I^{2}_{1}$ denotes the feature of $2^{nd}$ patch of the $1^{st}$ image, while $T_1$ refers to the text feature for class 1. Branch 1 (Top): A fusing module (blue) merges patch features into a single detail-enhanced image feature, which is then calculate similarity with the text feature (green). Branch 2 (Bottom): Patch features are individually compared to the text feature, and the maximum similarity is selected. The query proxy loss is then computed using the similarity distributions from both branches.}    
    \label{fig:loss}
    \end{center}
\end{figure}

We propose a fusion model and a proxy loss to merge multiple features from pre-generated image patches into a single, detail-rich vector. Our model uses image-agnostic text features, derived from class prompts, as a proxy to guide the fusion process and inject detailed information. The overall framework is depicted in Figure \ref{fig:loss}.

\subsubsection{Fusing Model}
The core principle is to combine features from different patches into a single new feature while preserving as much detail as possible. We implement a fusing model using a small transformer. As illustrated in Figure \ref{fig:loss}, this model takes the set of patch features as its input source and the original whole-image feature as its input target, along with image-agnostic class-level text features.

\subsubsection{Query Proxy Loss}
For small objects within an image, there will be one patch feature $u_{\text{max}} \in \mathbb{R}^{d}$ containing the most information about the object. 
The loss function aims to inject the most detailed patch feature into the fused feature.
We use a text feature $w \in \mathbb{R}^{d}$ of the description of the small object as a proxy to align the fused feature $v \in \mathbb{R}^{d}$ with the patch feature $u_{\text{max}}$ in their text-image joint embedding space. 
The loss minimizes the distance between the two similarity scores, $\textit{sim}(v, w)$ (the similarity of the fused feature to the proxy feature) and $\textit{sim}(w, u_{\text{max}})$ (the similarity of the top patch to the proxy feature). We aim to learn $v$ by minimizing the distance:
\begin{equation}
    \mathbf{L_{qp}} = \mathbf{D}[\textit{sim}(v, w), \textit{sim}(w, u_{\text{max}})]
\end{equation}
\noindent where $\mathbf{D}$ is the $L_2$ distance and \textit{sim} is the cosine similarity. The method requires dataset-level supervision (class names in the dataset), not fine-grained annotations. 
This loss is well-suited for real-world retrieval tasks (use a text query to retrieve all images containing certain small objects), where the presence of an object matters more than its instance count. The use of $L_2$ distance ensures that non-maximal patches still contribute to the gradient, while our fusion model is designed to integrate information across patches. The method can handle multiple instances within the same patch.

\subsection{Pseudo Code of the Algorithm}

\begin{lstlisting}[language=Python, caption=Pytorch-Like Code, label=loss_code]
from einops import rearrange


def compute_loss(fusing_model, patch_feature, vanilla_feature, text_feature):
    # b: batch size
    # p: number of patch
    # t: number of text feature
    # f: feature dim
    # vanilla_feature: clip feature for entire image, (b, f)
    # patch_feature: clip feature for different patches, (b, p, f)
    # text_feature: clip feature for text prompts, (t, f)
    # fusing_model: DetailCLIP model
    # All features are normalized.
    DetailCLIP_feature = fusing_model(patch_feature, vanilla_feature)
    patch_feature = rearrange(patch_feature, 'b p f -> (b p) f')
    proxy_feature = text_feature
    # (t,  f) @ (f, b * p) -> (t, b * p)
    q_p_similarity = proxy_feature @ patch_feature.T
    q_p_similarity = rearrange(q_p_similarity, 'k (b p) -> k b p')
    # (t, b, p) -> (t, b)
    q_p_similarity_max = q_p_similarity.max(-1)
    # (t, f) @ (f, b) -> (t, b)
    q_c_similarity = proxy_feature @ DetailCLIP_feature.T
    query_proxy_loss = mse_loss(q_p_similarity_max, q_c_similarity)
\end{lstlisting}


\section{Benchmark} 
\label{benchmark}
In text-image retrieval, the traditional task of "using a caption to retrieve a single image" is widely employed to evaluate a model's retrieval capability. However, the practical requirement of "using a word to retrieve all related images in a database" remains unaddressed. Traditional text-image retrieval datasets like Flickr30k and COCO-caption (MSCOCO dataset utilizing captions for retrieval) are not suitable for evaluating the latter task. 

\begin{figure}[ht]
    \includegraphics[width=0.5\textwidth]{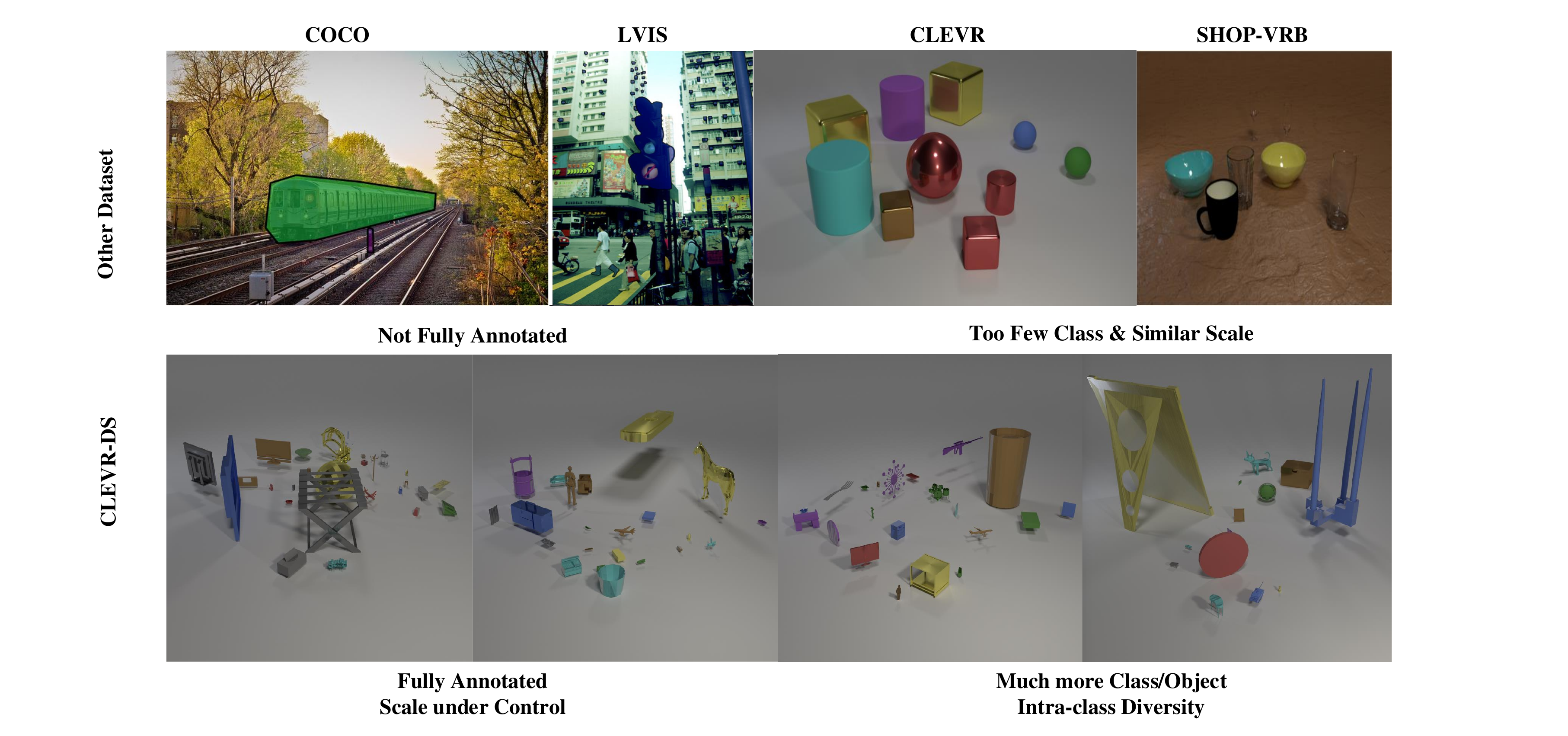}
    \caption{Illustration for different datasets. Existing datasets have different flaws for the retrieval by class name task. COCO and LVIS are not fully annotated, while CLEVR and SHOP-VRB, though fully annotated, have a limited number of classes and monotonous object scales. In contrast, our proposed CLEVR-DS dataset is fully annotated for retrieval tasks and features a greater variety of classes and objects with different scales, thereby increasing intra-class diversity.}
    \label{fig:OtherDataset}
\end{figure}

In theory, numerous existing datasets with class-wise object supervision and class names can be utilized to construct the benchmark. However, both have deficits not only for the proposed task but also for analyzing methods' detail retrieval capabilities. Examples are displayed in Figure \ref{fig:OtherDataset}.

For real-world datasets, the label information in Visual Genome \cite{krishna2017visual}, ImageNet-1K \cite{imagenet}, and GPR1200 \cite{gpr1200} primarily focuses on the main object in the image. As these datasets are sourced from real-world scenes, achieving complete-annotation of the images is challenging, resulting in insufficient fine-detail annotation. Furthermore, the large dataset LVIS \cite{gupta2019lvis} contains many missing labels, making it impossible to draw accurate conclusions during retrieval evaluation. Synthetic datasets such as CLEVR and SHOP-VRB by \cite{SHOP-VRB} have too few object categories. Additionally, these datasets lack specific design considerations for object size in the images.

\subsection{CLEVR-DS}
To achieve accurate retrieval evaluation, we created a dataset called "CLEVR of Different Scales (\textbf{CLEVR-DS})," which includes 138 categories from ShapeNet and has an average of nearly 14 instances per image, surpassing LVIS. This dataset allows for complete annotation information (e.g., spatial position, bounding box, category, and attribute) in the image. Our CLEVR-DS covers a wide range of object scales (small, medium, and large) and exhibits greater variability in scene complexity. 

\subsection{Data Statistics of CLEVR-DS} 
\label{datastatistics}
\begin{table}[htbp]
  \centering
  \caption{Detailed statistics of datasets used in our experiments.}
  \label{table:dataset_stats}
  \small
  \begin{tabular}{lcccc}
    \toprule
    \textbf{Dataset} & \textbf{Per Image} & \textbf{Mean} & \textbf{Min} & \textbf{Max} \\
    \midrule
    \multirow{2}{*}{CLEVR-DS} & Instance & 13.78 & 1 & 50 \\
                               & Class    & 13.78 & 1 & 50 \\
    \midrule
    \multirow{2}{*}{MSCOCO}    & Instance & 7.33  & 1 & 93 \\
                               & Class    & 2.92  & 1 & 18 \\
    \midrule
    \multirow{2}{*}{Unity-Retail} & Instance & 25.52 & 16 & 42 \\
                                  & Class    & 13.52 & 9  & 19 \\
    \midrule
    \multirow{2}{*}{LVIS}      & Instance & 11.2  & 1 & 294 \\
                               & Class    & 3.4   & 1 & 24  \\
    \bottomrule
  \end{tabular}
\end{table}

\begin{table}[htbp]
    \begin{center}
    \small
    \begin{tabular}{ccccc}
        \toprule
        DATASET & COCO  & LVIS & CLEVR-DS & Unity-Retail\\
        \midrule
        Image Number & 122,219  & 122,219 & 10,000 &1,000\\
        Class Number & 80  & 1230 & 138   &16 \\
        \bottomrule
    \end{tabular}
    \end{center}

    \caption{Overall Statistics of datasets.}
        \label{table:classnumber}
\end{table}

\begin{figure}[ht]
    \centering
    \includegraphics[width=0.4\textwidth]{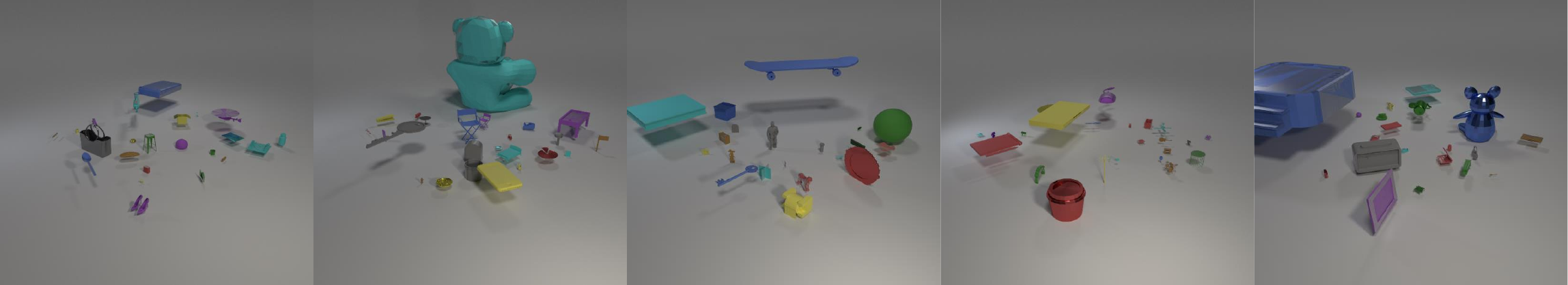}
     \caption{Our synthetic CLEVR-DS dataset illustrated above, is similar to the original CLEVR dataset but presents more complex scenes. We randomly scatter 1-50 instances from 138 ShapeNet object classes within each image. These images feature 1-3 large objects and more than 10 small objects, significantly increasing the challenge for the text-to-image retrieval task.}
    \label{fig:clevrvis_new}
\end{figure}
We have selected several datasets for performing the class with prompt text-to-image retrieval tasks. The statistics for CLEVR-DS, Unity-Retail, MSCOCO, and LVIS can be found in Table \ref{table:dataset_stats} and Table \ref{table:classnumber}. These four datasets contain relatively more instances per image compared to other datasets, increasing the likelihood of small objects within the images. Notably, the mean number of instances per image for CLEVR-DS is higher than that of LVIS. A demo for CLEVR-DS dataset is shown in Figure ~\ref{fig:clevrvis_new}.

\section{Experiments} 
\label{experiment}

\subsection{Implementation Details}
\subsubsection{Dataset setting} 
We evaluate our model on both synthetic and real-world datasets. On our fully annotated synthetic dataset, \textbf{CLEVR-DS}, we test model's image fine-detail retrieval ability by setting the size of objects in the image: \textbf{CLEVR-DS-S} and \textbf{CLEVR-DS-L}, which contain only small or large objects with respect to the query.
For MSCOCO, we randomly select 5000 images for validation and use the original validation set for testing. LVIS shares the same images as MSCOCO and we apply the same split setting as MSCOCO.
\subsubsection{Model architecture}
We build our framework on three different CLIP-like encoders: two ViT-B/16 models from SLIP and the original ViT-L/14 from CLIP. Feature dimensions are 512 or 768, depending on the backbone. For patch and text feature extraction, we use the original, unmodified image and text encoder. Our fusion model is a small transformer (3 encoder and 3 decoder layers), and the entire framework is optimized using query proxy loss.
\subsubsection{Patch selection method}
We use our CC method, specifically \textbf{CC}@10, which generates 166 patches per image with 10 as the scale sensitivity hyper-parameter. We evaluate the CLIP-like model's retrieval ability for untrained retrieval using both the \textbf{CC} patch features (an image's score is its highest \textbf{CC} patch score.) and whole image features. In the DetailCLIP scenario, the fusion model takes the CC patch features as input and outputs a single feature.
\noindent \textbf{Training Details}. We train DetailCLIP for 10 epochs on a single GTX 2080ti with a batch size of 30. The model is optimized using AdamW \cite{loshchilov2019decoupledweightdecayregularization} with a linear learning rate scheduler and warm-up. To accelerate training, we pre-extract all patch features offline in parallel using Ray\footnote{https://docs.ray.io/en/master/}. Hyperparameters are tuned on a validation set, and final results are reported on a held-out test set.
\subsubsection{Evaluation Metrics}
\label{Evaluation Metrics} 
The current evaluation metric for text-to-image retrieval calculates the top $k$ recall accuracy of a text query, which is insufficient for our goal of retrieving \textbf{all} related images in a database, therefore we propose a new metrics:
\begin{equation}
\text{Recall@k} = \frac{t_k}{n}
\end{equation}
where $n$ is the total number of ground-truth images for a given query class, and $t_k$ is the number of correct images found within the top $n \times k$ retrieved results. We report Recall@1, Recall@3, and Recall@5 for all evaluations.

\subsection{Main Result Analysis}
We clarify two approaches for \textbf{CC}@10 inputs. Our \textbf{DetailCLIP} uses a \textbf{"single feature"} approach, where all patch features are fused into one vector. In contrast, \textbf{CC}@10 input with "non-single features" indicates the use of multiple image feature vectors (one for each patch), which is computationally expensive and inconvenient for end-to-end retrieval tasks or other downstream tasks. Theoretically, both \textbf{CC}@10 input contains an equal amount of detailed information. As a result, \textbf{CLIP with cc@10 (multiple features) should not only serve as a baseline but also be comparable to DetailCLIP with cc@10 input} in terms of retrieval performance.

\subsubsection{Results on CLEVR-DS with Controlled Object Sizes}
Using a CLIP ViT-B/32 backbone and a small 6-layer transformer for feature fusion, we evaluate retrieval on \textbf{CLEVR-DS}. In non-training setup, we test two retrieval methods: single-feature, which uses the text to retrieve whole image's feature, and multi-feature, which uses text to retrieve patch features, and use that patch to represent the corresponding image.
As shown in Table \ref{table:MainResult}, on the mixed-size CLEVR-DS dataset, \textbf{DetailCLIP} improves retrieval performance from 10.61\% to 22.54\%. The improvement is most significant on the CLEVR-DS-S (small object) subset with a 10.23\% gain, and a 2.76\% gain is observed on the CLEVR-DS-L (large object) subset.

\begin{table}[htbp]
  \resizebox{\columnwidth}{!}{
  \begin{tabular}{ccccccccc}
    \toprule
    Dataset & Method & Single Feature & Input & Recall@1 & Recall@3 & Recall@5 \\ 
    \midrule
    ~ & CLIP & \ding{52} & Full Image & 10.61\% & 29.64\% & 49.50\% \\
    CLEVR-DS & CLIP & \ding{53} & \textbf{CC}@10 & \textbf{22.56}\% & 45.45\% & 64.91\% \\
    ~ & \textbf{DetailCLIP} &\ding{52}& \textbf{CC}@10 & 22.54\% & \textbf{46.16}\% & \textbf{64.98}\% \\ 
     \midrule
    ~ & CLIP &\ding{52}& Full Image & 4.43\% & 16.25\% & 30.32\% \\ 
    CLEVR-DS-S & CLIP&\ding{53} & \textbf{CC}@10 & 14.57\% & 32.55\% & \textbf{37.12}\% \\
    ~ & \textbf{DetailCLIP} &\ding{53}& \textbf{CC}@10 & \textbf{14.66}\% & \textbf{32.77}\% & 32.78\% \\ 
     \midrule
    ~ & CLIP&\ding{52} & Full Image & 13.57\% & 23.19\% & 28.65\% \\ 
    CLEVR-DS-L & CLIP&\ding{53} & \textbf{CC}@10 & 15.30\% & \textbf{25.93}\% & \textbf{31.97}\% \\
    ~ & \textbf{DetailCLIP}&\ding{52} & \textbf{CC}@10 &\textbf{16.33}\% & 25.47\% & 30.96\% \\
     \bottomrule
\end{tabular}
}
  \caption{Performance of DetailCLIP and CLIP in CLEVR-DS.}
  
  \label{table:MainResult}
\end{table}

\subsubsection{Results on Commonly Used Dataset}
We compare DetailCLIP against several VLMs. As the class-wise retrieval task in this paper differs slightly from the COCO-style caption retrieval task, the results from their original papers are not directly comparable. We test against various CLIP backbones (ViT-B/16, ViT-L/14) and other leading approaches, including detector-based models like RegionCLIP (ResNet50-C4), MDETR (EfficientNet-B5), and GLIP (Swin-L), as well as the CLIP variant SLIP (ViT-B) and BLIP2 (ViT-G) and InternVL-C. We emphasize that DetailCLIP is a \textbf{detector-free} framework. This means the performance isn't limited by a detector, our \textbf{CC} scheme operates in constant time, and the patches inherently capture relationships between objects in multiple scales.

To confirm our method's effectiveness on complex scenes, we tested on the real-world MSCOCO and LVIS datasets and the fully-annotated synthetic Unity-Retail dataset. 
As shown in Table \ref{table:MainResult2}, DetailCLIP consistently outperforms the full image baseline and \textbf{CC}@10 baseline on most datasets. On MSCOCO, using ViT-L/14 backbone, it improves Recall@1 by $\sim$6\% (compared with full image baseline), demonstrating its superiority over the original CLIP feature. This performance gain is even larger on fully annotated synthetic datasets like Unity-Retail ($\sim$17\%) and CLEVR-DS ($\sim$20\%), suggesting that the performance gap is linked to annotation completeness. 
On MSCOCO, DetailCLIP's performance is competitive with competitive selective methods. On the fully annotated CLEVR-DS, it surpasses all compared approaches. 
The multi-feature CLIP + \textbf{CC}@10 serves as a performance upper bound (requiring 166 comparisons per image), and our method is far more efficient, retrieving with only a single feature per image.
Combined with the training-free \textbf{CC} and a light fusion module, even a shallow model like CLIP can outperforms the powerful models such as BLIP2 and InternVL, offering a cost-effective solution for large-scale retrieval tasks.
\begin{table*}[htbp]
    \resizebox{2\columnwidth}{!}{
    \begin{tabular}{ccccccccc}
        \toprule
        \multicolumn{3}{c}{\textbf{DATASET}} &{\textbf{CLEVR-DS}}& \textbf{LVIS} &\textbf{COCO}&\textbf{Unity} \\ \midrule
         Method & Single Feature & Input  & Recall@1 & Recall@1 & Recall@1& Recall@1 \\
        \midrule
          CLIP-ViT-B/16  & \ding{52} & Full Image &8.51\% & 7.49\% & 40.93\% & 24.63\% \\
          CLIP-ViT-B/16 & \ding{53} & \textbf{CC}@10 &17.03\% & \textbf{9.40}\% & 41.24\% & 23.11\%\\
        \textbf{DetailCLIP-ViT-B/16} & \ding{52}& \textbf{CC}@10 &\textbf{18.09}\% & 7.66\% & \textbf{44.19}\% & \textbf{25.02}\% \\ \midrule
         CLIP-ViT-L/14  & \ding{52}& Full Image  &13.81\% & 15.12\% & 56.74\% & 35.74\%\\ 
         CLIP-ViT-L/14 & \ding{53}& \textbf{CC}@10  &33.21\% & \textbf{22.00}\% & 59.40\% & 52.40\%\\
        \textbf{DetailCLIP-ViT-L/14} & \ding{52}& \textbf{CC}@10 &\textbf{33.46}\% &15.29\% & \textbf{62.63}\% & \textbf{55.21}\%\\
        \midrule
         RegionCLIP (detector-based)  & \ding{52}& Full Image & 10.98\% & 10.13\% & 46.06\% & 24.10\% \\
        MDETR (detector/grounding-based) & \ding{52}& Full Image & 9.82\% & 12.24\% & 49.65\% & 25.75\% \\
        GLIP (detector/grounding-based) & \ding{52}& Full Image & 11.84\% & 13.75\% & 50.87\% & 28.31\% \\
         SLIP (CLIP-like) & \ding{52}& Full Image & 9.51\% & 9.65\% & 47.49\% & 24.42\% \\
        BLIP2 (CLIP-like)& \ding{52}& Full Image & 17.14\% & 13.36\% & 47.85\% & 32.34\% \\
        InternVL-C (MLLM) & \ding{52}& Full Image & 22.45\% & 16.81\% & 57.12\% & 38.93\% \\
        \bottomrule
    \end{tabular}
    }
  \caption{Performance of DetailCLIP framework and other methods}
  \label{table:MainResult2}
\end{table*}

\subsection{DetailCLIP used as Plug-in Module for Other VLMs}
\begin{table}[htbp]
    \resizebox{\columnwidth}{!}{
    \begin{tabular}{cccc}
        \toprule
        Method & CLIP-ViT-B/16 & SLIP-ViT-B/16  & BLIP2\\ 
        \midrule
        Vanlilla & 8.51\% & 9.51\%  & 17.14\% \\ 
        Detail-X & 18.09\% & 19.39\% & 29.89\% \\
        \bottomrule
    \end{tabular}
    }
    \caption{Detail-X plug-in module for different VLMs.}
    \label{table:plug_in}
\end{table}

We tested our DetailCLIP framework as a plug-in module, termed "\textbf{Detail-X}," on the CLEVR-DS dataset.
Integrating Detail-X with SLIP and BLIP2 (creating DetailSLIP and DetailBLIP2) significantly improved recall@1 scores by 9\% to 13\% (Table \ref{table:plug_in}).

\section{Ablation Study}
We analyze our framework, data, and retrieval methods to further showcase the effectiveness of the method proposed in the paper and its components. In all experiments from this section, we use CLIP-ViT-B/32 as the default feature
extractor if not mentioned specifically.

\subsection{Patch Generation and Upper Bound Analysis}
\begin{figure}[htb]
    \centering
    \includegraphics[width=0.21\textwidth]{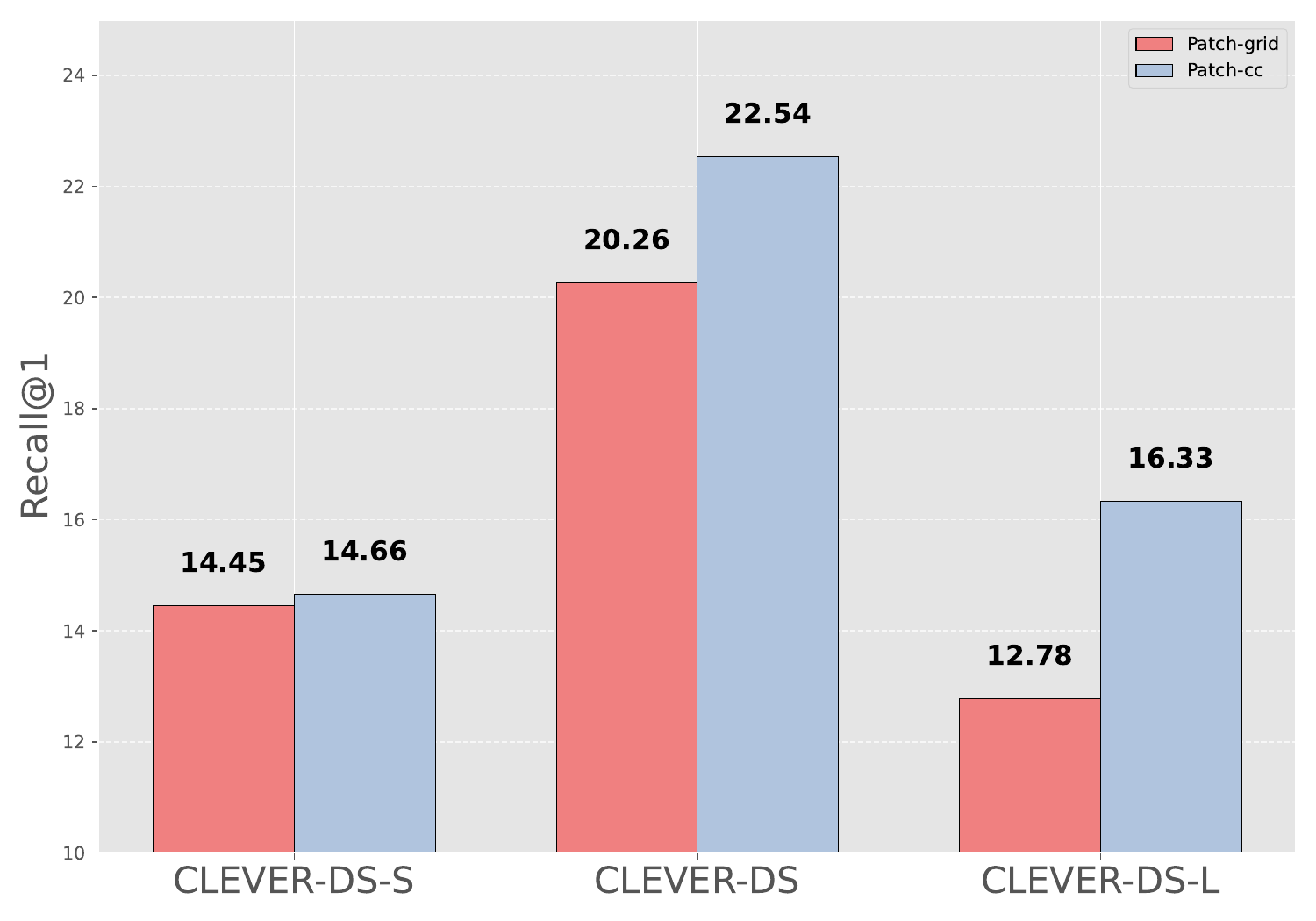}
    \includegraphics[width=0.26\textwidth]{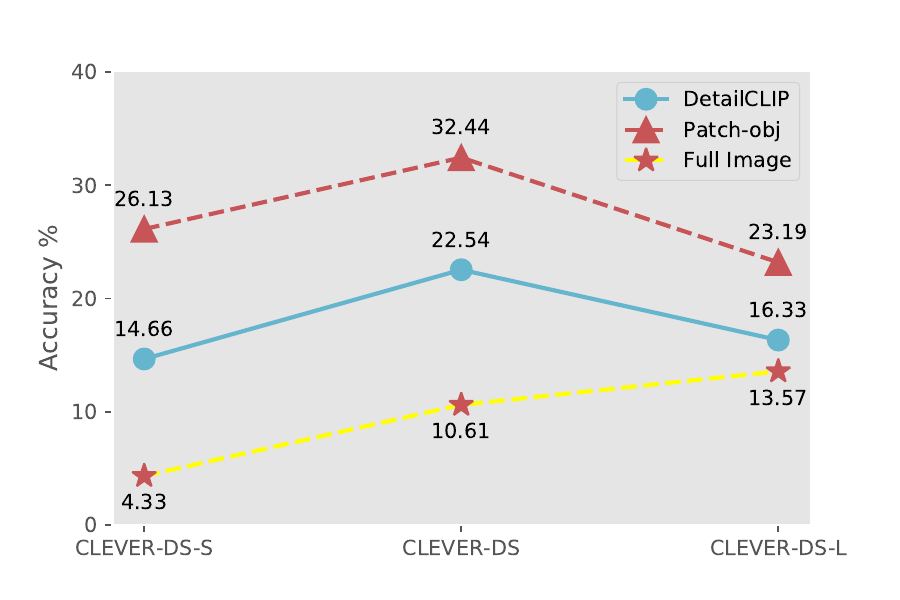}
    \caption{Retrieval performance under different patch generation schemes and upper bound analysis using ground-truth annotation.}
    
    \label{fig:grid-cc}
\end{figure}

First, we compare two patch generation schemes on CLEVR-DS: a simple non-overlapping "Patch-grid" (ViT style) and our "Patch-cc" method (Complete Cover). As shown in Figure \ref{fig:grid-cc} (left), Patch-cc consistently outperforms Patch-grid across all object sizes on the CLEVR-DS dataset, confirming its superior ability to generate more effective patches with varying levels of detail for fusion.

Second, to establish a theoretical performance upper bound, we created a "Patch-obj" method. It simulates a perfect detector by cropping objects using their ground-truth bounding boxes to generate the patches. As shown in Figure \ref{fig:grid-cc} (right), our DetailCLIP method (22.54\%) approaches the performance of this "Patch-obj" upper bound (32.44\%) more closely than the standard full-image baseline.

\subsection{Different Fusion Approaches}

Our method has been trained on the target dataset in an unsupervised manner. To evaluate the method's effectiveness, we experiment with 4 different adaptation methods on the CLEVR-DS dataset that each adds a trainable module on top of the CLIP like our approach. We tested two types of modules, MLPs and Transformer, which use the original CLIP feature as input and the adapted feature on the target domain as output. Results in Table \ref{fintune} show that the vanilla adaptation on the target dataset is not as effective as our method.

\begin{table}[htbp]
    \small
    \begin{center}
    \scalebox{1.0}{
    \begin{tabular}{|cc|cc|}
    \hline
        \multicolumn{2}{|c|}{Module} & \multirow{2}{*}{Patch-cc} & \multirow{2}{*}{Recall@1} \\ \cline{1-2}
        Transformer& \# of MLP-layers & &\\ \hline
        \ding{53} & 0        &\ding{53} &10.61\%\\ 
        \ding{53} & 1        &\ding{53}  &15.91\% \\ 
        \ding{53} & 2        &\ding{53}   &14.70\% \\ 
        \ding{53} & 3        &\ding{53}    &13.96\% \\ 
        \ding{52} & \ding{53} & \ding{53} &15.90\%\\ 
        \ding{53} & 3        &\ding{52}    &17.21\%\\ 
        \ding{53} & 1        &\ding{52}    &17.51\%\\ 
        \ding{52} & \ding{53} &\ding{52}&\textbf{22.54}\%\\ \hline
    \end{tabular}
    }
    \end{center}
    \caption{Add the trainable modules on top of the CLIP model to fine-tune}
    \label{fintune}

\end{table}




\subsection{Ablation on different object numbers}

The number of instances and categories in the evaluation dataset is crucial to the performance of the image retrieval method. As shown in Table~\ref{table:diff_object_num}, we tested our method on a dataset with fewer instances per image and fewer categories. The results show that our method achieves better performance with fewer objects in an image.

\begin{table}[htb]

    \resizebox{0.98\columnwidth}{!}{
    \begin{tabular}{ccccc}
        \toprule
        Dataset & categories  & instance per image  & Recall@1 & Recall@5 \\ 
        \midrule
        CLEVR-DS & 138 &14  & 22.54\%  & 64.98\% \\ 
        CLEVR-DS & 51  &3 &  41.10\% & 73.41\% \\
        \bottomrule
    \end{tabular}
    }
    \caption{Retrieval Performance of DetailCLIP and CLIP on CLEVR-DS with different object numbers.}
    \label{table:diff_object_num}
\end{table}

\subsection{DetailCLIP Performance Under Different Complete Cover Schemes} \label{diff_ccs}

We present the number of patches at different levels based on varying $c$ values in Table \ref{table:cc_structure}. For notational convenience, we set $\frac{a}{c} = k$.
\begin{table}[ht]
    \small
    \begin{center}
    \scalebox{0.6}{
    \begin{tabular}{|c|c|c|c|c|c|c|c|c|}
    \hline
        CC@k & Patch Numbers& Level 1 & Level 2 & Level 3 & Level 4 & Level 5 & Level 6 & Level 7 \\ \hline
        1 & 1 & 1 & ~ & ~ & ~ & ~ & ~ & ~ \\ \hline
        2 & 5 & 1 & 4 & ~ & ~ & ~ & ~ & ~ \\ \hline
        3 & 14 & 1 & 4 & 9 & ~ & ~ & ~ & ~ \\ \hline
        4 & 30 & 1 & 4 & 9 & 16 & ~ & ~ & ~ \\ \hline
        5 & 39 & 1 & 4 & 9 & 25 & ~ & ~ & ~ \\ \hline
        6 & 66 & 1 & 4 & 9 & 16 & 36 & ~ & ~ \\ \hline
        7 & 79 & 1 & 4 & 9 & 16 & 49 & ~ & ~ \\ \hline
        8 & 103 & 1 & 4 & 9 & 25 & 64 & ~ & ~ \\ \hline
        9 & 136 & 1 & 4 & 9 & 16 & 25 & 81 & ~ \\ \hline
        10 & 166 & 1 & 4 & 9 & 16 & 36 & 100 & ~ \\ \hline
        11 & 187 & 1 & 4 & 9 & 16 & 36 & 121 & ~ \\ \hline
        12 & 248 & 1 & 4 & 9 & 16 & 25 & 49 & 144 \\ \hline
        13 & 273 & 1 & 4 & 9 & 16 & 25 & 49 & 169 \\ \hline
        14 & 315 & 1 & 4 & 9 & 16 & 25 & 64 & 196 \\ \hline
        15 & 355 & 1 & 4 & 9 & 16 & 36 & 64 & 225 \\ \hline
    \end{tabular}
    }
    \end{center}
    \caption{Number of patches at different levels.}
    \label{table:cc_structure}
\end{table}

\begin{figure}[ht]
    \centering
    \includegraphics[width=0.5\textwidth]{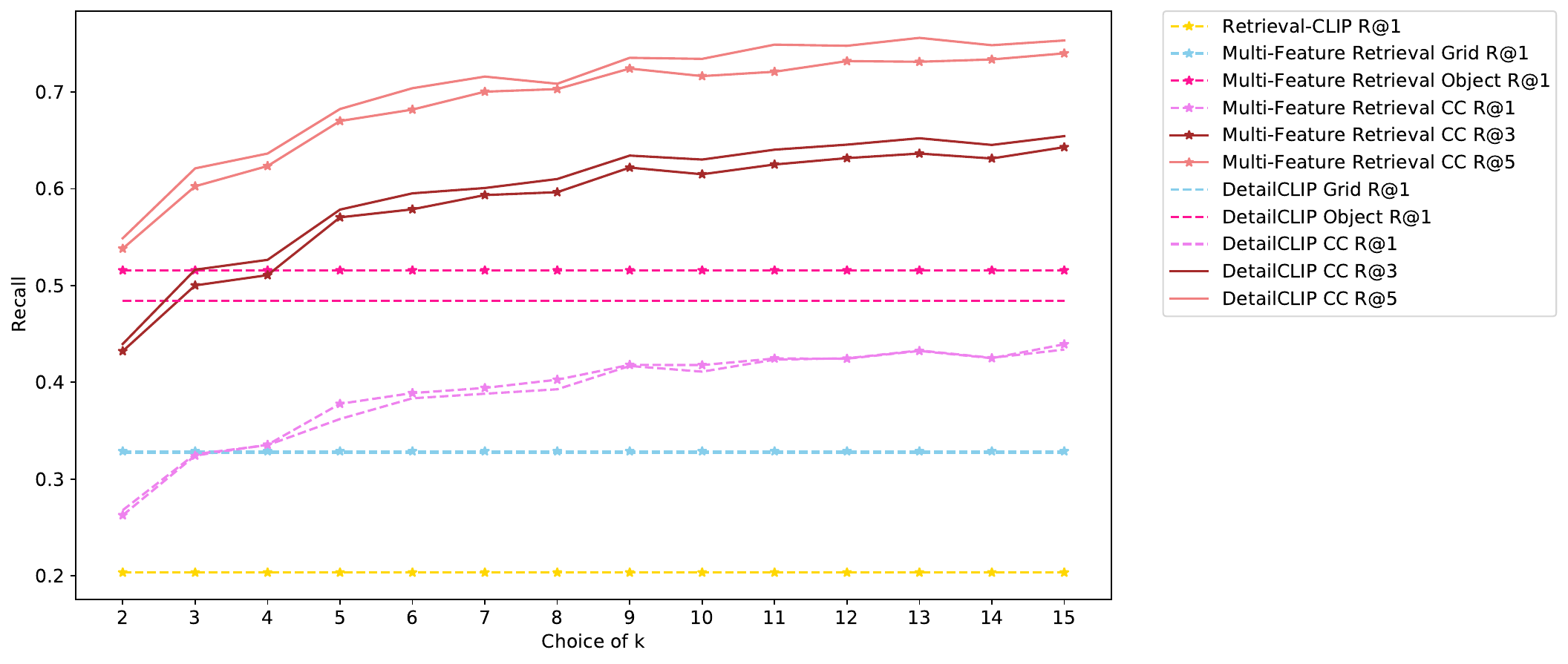}
    \caption{Recall for retrieval and DetailCLIP models with different $k$ Values}
    \label{fig:cc_recall}
\end{figure}

To determine an appropriate $k$, we examine the performance of the DetailCLIP model across 14 distinct $k$ values using the CLEVR-DS dataset (51 classes version). The results are displayed in Figure~\ref{fig:cc_recall}. Lines with star markers represent retrieval results, while those without indicate DetailCLIP results. Dashed lines correspond to recall@1 results, and the same color is used for the same patch selection method under identical recall metrics. The figure reveals several observations:

\begin{itemize}
\item The recall@1 results for DetailCLIP (single feature) are on par with the retrieval baseline (multi-feature) for any $k$. DetailCLIP performs marginally better with larger $k$ values.
\item The performance of both DetailCLIP and the retrieval baseline starts to increase at $k = 2$ and plateaus at $k=9$. For recall@1, patch-cc outperforms patch-grid beginning at $k=4$.
\item For $k \geq 8$, the number of patches grows significantly, but the performance of the DetailCLIP model does not.
\end{itemize}

Based on the above analysis and considering the trade-off between DetailCLIP performance and computational complexity, we opt for $k=10$ for the experiments presented in the main body of our paper. Notably, when $k=10$, the last level of patch number aligns with the patch-grid method.

\subsection{Information Injection Ablation}

\begin{figure}[ht]
	\centering
	\includegraphics[scale=0.5]{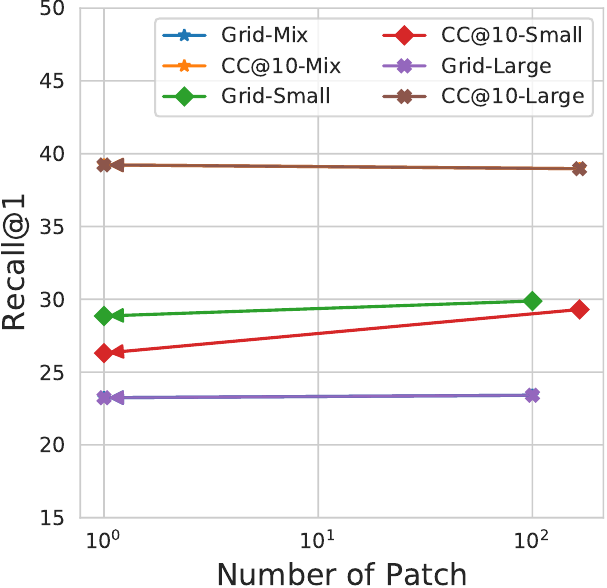}
	\caption{After applying DetailCLIP, we achieve 100x \#Patch Reduction with no significant performance loss.}
	\label{fig:DetailCLIP_ablation}
	
\end{figure}

We test the ability of the DetailCLIP framework to inject detailed information using a varying number of patches. The CLEVR-DS dataset (51 classes version) is used for the task. We train the DetailCLIP fusing model with different numbers of patches and test the retrieval performance of the fused feature. In Figure \ref{fig:DetailCLIP_ablation}, "Grid" and "CC" represent the patch selection methods, while "Mix," "Small," and "Large" indicate the object scale in the CLEVR-DS dataset.

As shown in Figure \ref{fig:DetailCLIP_ablation}, DetailCLIP's performance remains stable as the number of patches increases, demonstrating that our framework can effectively inject detailed information from multiple patches. The results also indicate that the patch generation method, whether patch-cc or patch-grid, yields similar results for DetailCLIP, comparable to the multi-feature CLIP results. This further proves that DetailCLIP can effectively inject information from various sources. This figure also shows a trend that moving from large to small objects increases the recall of DetailCLIP, whereas the recall decreases for the CC method. The reason is patch-grid can only cover small objects, but patch-cc can cover objects on any scale.

\section{Hyper-parameter Tuning}
\label{hyperparam}

For training DetailCLIP, we employ the AdamW optimizer, a linear learning rate scheduler, and a linear warm-up training strategy over 10 epochs. DetailCLIP is trained using a single GTX 2080ti in 1-2 hours (the vanilla CLIP features are pre-extracted and shared across different experiments), with a batch size of 30 across all experiments.

For different $k$ values, we independently perform a grid search of DetailCLIP hyperparameters, as shown in the table \ref{table:hyper}. We select the optimal hyperparameters on a validation set and report the results on a held-out test set.

\begin{table}[ht]
    \centering
    \resizebox{0.98\columnwidth}{!}{
    \begin{tabular}{|c|c|}
    \hline
        Name&Candidate  \\ \hline
        Learning Rate &  [0.001, 0.003, 0.005, 0.007, 0.01]   \\  \hline
        Weight Decay&  [0, 0.001]    \\ \hline
        Step Size for Learning Rate Decay &  [60, 120]    \\ \hline
        Gamma Value for Learning Rate Decay& [0.5, 0.7, 0.9]      \\ \hline
        Gradient Clip Value&  [0.00001, 0.0001]      \\ \hline
        Layer Normalization’s Epsilon & [0.0001, 0.001, 0.01]    \\ \hline
    \end{tabular}
    }
    \caption{Hyper-Parameter candidate in validation Set.}
    \label{table:hyper}
\end{table}

\section{Computational Resource and Efficiency Analysis} 
\label{time}

As demonstrated in Table~\ref{table:time}, the inference time has been reduced from 4.9ms to 1.3ms, which is a reduction of around 70\%. This significant improvement can potentially increase the speed of processing data, making retrieval tasks more efficient and responsive. Faster inference time can also lead to an improved user experience, especially in real-time scenarios where speed is critical.

Additionally, the reduction in memory usage from 333KB to 2.2KB is impressive. This represents a reduction of over 99\%, which can be crucial for image retrieval tasks with limited resources. Lower memory usage means that the system can run on lower-end devices, which can result in cost savings for the hardware used to run the system.

Moreover, only the forward pass is needed (inference) when extracting the CLIP feature, and the text feature of class names in the dataset can be extracted once for all. A tiny fusion model is the only trainable part. \textbf{The FLOPs for the fusion model is 0.013G, compared with CLIP's 13G}. We strictly implement the CC algo following Section \ref{completecover} to guarantee objects in any scale are covered.

\begin{table}[htb]

    \resizebox{0.98\columnwidth}{!}{
    \begin{tabular}{ccccc}
        \toprule
        Method & patches  & time(ms)  & Memory(KB) & Recall@1 \\ 
        \midrule
        CLIP + CC@10 & 166 & 4.9  & 333  & 14.57\% \\ 
        \textbf{Detailclip + CC@10} & 1 & \textbf{1.3}  &\textbf{2.2} & \textbf{14.66}\% \\
        \bottomrule
    \end{tabular}
    }

    \caption{Computational Resource}
    \label{table:time}
\end{table}

\section{Conclusion}

In conclusion, we present DetailCLIP, a detail-injected feature fusion framework for detailed text-to-image retrieval. By combining our \textbf{Complete Cover} patching scheme with a Transformer-based \textbf{fusion model} and a \textbf{query proxy loss}, DetailCLIP produces a single, detail-rich feature representation. To validate our approach, we built the \textbf{CLEVR-DS} benchmark. Extensive experiments show that DetailCLIP significantly outperforms standard CLIP-like models. Furthermore, our framework can serves as an effective \textbf{plug-in module} to enhance the detail retrieval capabilities of other VLMs.

\bibliographystyle{IEEEbib}
\bibliography{strings,refs}

\end{document}